\documentclass[letterpaper, 10 pt, conference]{ieeeconf}  

\IEEEoverridecommandlockouts                              

\usepackage{amssymb}
\usepackage{bbm}
\usepackage{graphicx}
\usepackage{amsmath}
\usepackage{amssymb}
\usepackage{subfig}
\usepackage{tabularx}
\usepackage{slashbox}
\usepackage{graphicx,wrapfig}

\usepackage[nolist]{acronym}
\usepackage[textsize=small]{todonotes}

\long\def\gd#1{\bf {\color{red}Greg: $1}}
\long\def\gd#1{\bf {\color{blue}Lucas: $1}}

\long\def\invis#1{}
\usepackage[normalem]{ulem}

\newcommand\gderror[1]{
  \typeout{--------------------------------------------------------------------}
  \typeout{------- #1 ---------}
  \typeout{--------------------------------------------------------------------}
  {\bf #1}
}
\newcounter{gdTmp}
\newcounter{gdLastCount}
\newcommand\maxpage[2][Error]{  
\ifnum\value{page}>#2
   \gderror{\Large \bf On page {\thepage} we are past page #2 (too long).   #1 }
\else\fi
}
\newcommand\maxpageSinceLast[2][Error]{  
\ifnum \numexpr \value{page} - \value{gdLastCount}\relax>#2
   \gderror{Exceeds max length #2 pages. Page \thepage: #1}
\thepage\else\fi
\setcounter{gdLastCount}{\value{page}}
}

\long\def\gd#1{{{\bf\color{red} Greg: #1}}}

\maxpage[{\color{red}\Large Main text is too long.}]{6}

\usepackage{caption}
\usepackage{hyperref}
\usepackage{amsmath, amsfonts}
\usepackage{dsfont}

\def\andrew#1{} 

\title{\LARGE \bf
	Uncertainty-aware hybrid paradigm of nonlinear MPC and model-based RL for offroad navigation: Exploration of transformers in the predictive model      
}

\author{Faraz Lotfi, Khalil Virji, Farnoosh Faraji, Lucas Berry, Andrew Holliday, David Meger, and Gregory Dudek$^{1}$
	\thanks{$^{1}$Authors are with the Mobile Robotics Lab (MRL), Faculty of Computer Science, McGill University,
		Montreal, Canada
		{\tt\small F.Lotfi@CIM.McGill.ca}}%
}

\begin{document}

	\maketitle
	\thispagestyle{empty}
	\pagestyle{empty}

	\begin{abstract} 
In this paper, we investigate a hybrid scheme that combines nonlinear model predictive control (MPC) and model-based reinforcement learning (RL) for navigation planning of an autonomous model car across offroad, unstructured terrains without relying on predefined maps. Our innovative approach takes inspiration from BADGR, an LSTM-based network that primarily concentrates on environment modeling, but distinguishes itself by substituting LSTM modules with transformers to greatly elevate the performance our model. Addressing uncertainty within the system, we train an ensemble of predictive models and estimate the mutual information between model weights and outputs, facilitating dynamic horizon planning through the introduction of variable speeds. Further enhancing our methodology, we incorporate a nonlinear MPC controller that accounts for the intricacies of the vehicle's model and states. The model-based RL facet produces steering angles and quantifies inherent uncertainty. At the same time, the nonlinear MPC suggests optimal throttle settings, striking a balance between goal attainment speed and managing model uncertainty influenced by velocity. 
In the conducted studies, our approach excels over the existing baseline by consistently achieving higher metric values in predicting future events and seamlessly integrating the vehicle's kinematic model for enhanced decision-making. 
The code and the evaluation data are available at (\href{https://github.com/FARAZLOTFI/}{Github-repo}).
     \end{abstract}


\section{Introduction}
Researchers in the field of off-road autonomous navigation have developed robust techniques to effectively drive vehicles toward their destinations. A key strategy involves the use of image-based predictive models to enhance planning and optimize reward functions~\cite{guastella2020learning, islam2022off}. A notable contribution in this area is BADGR \cite{badgr}, which combines model-based and model-free RL, effectively addressing path planning on smooth terrains while minimizing collision risks. BADGR prescribes actions based on image data and leverages predictive models to anticipate future event sequences. Subsequent researchers have harnessed diverse data sources and devised architectural enhancements. 
One notable example fuses drone aerial imagery with onboard visuals, improving perceptual capabilities~\cite{manderson2020-icra}, while another introduces a trajectory planner with constrained attention mechanisms integrated into the BADGR framework~\cite{wapnick2021}. 
However, these models lack the inclusion of uncertainty and focus solely on the robot's environment, neglecting its kinematics and dynamics. Addressing this issue would involve integrating motion constraints into planning and considering factors such as road friction, slope, and vehicle states, which are difficult to estimate from images alone.

\begin{wrapfigure}{l}{0.23\textwidth}
    \includegraphics[width=0.23\textwidth]{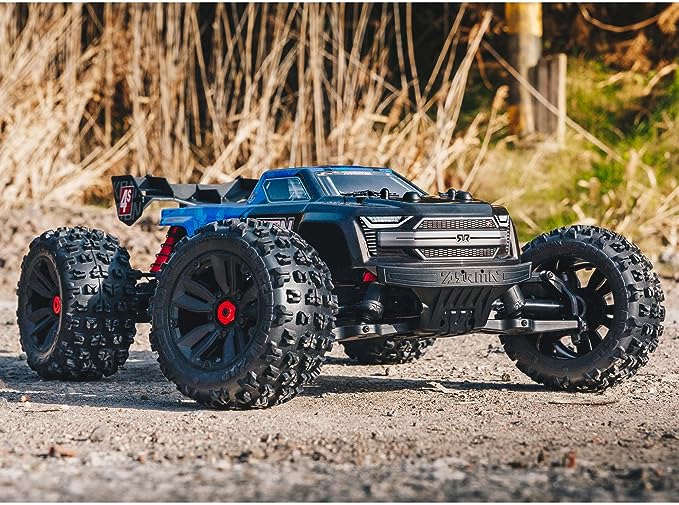}
    \caption{A model scale vehicle as used in our data collection.}
    \label{RC_car}
\end{wrapfigure}

The previously mentioned predictive models are related to model based control as used in control theory. Deep learning-based controllers excel at processing complex, high-dimensional input data, whereas MPC traditionally relies on a mathematical model to evaluate actions~\cite{garcia1989model}. 
MPC has seen extensive use in planning, particularly in navigation. An influential method in this domain is the Model-Predictive Path Integral (MPPI)~\cite{MPPI} approach, 
which introduces a stochastic optimal control framework. Numerous studies have been conducted using this technique, aimed at enhancing various facets of its application. For example, \cite{yin2023risk} incorporates Conditional Value-at-Risk (CVaR) to generate optimal control actions for safety-critical robotic applications, resulting in fewer collisions compared to the baseline MPPI. Additionally, \cite{sharma2023ramp} proposes a variable horizon planning MPPI, a topic further explored in the subsequent literature section on uncertainty incorporation. However, these approaches predominantly assume a static world when planning and neglect to model the dynamics of the environment based on the given set of actions. In other words, they tackle the planning problem by predominantly focusing on the vehicle's dynamics while inadequately representing the environment's dynamics.

Regarding the image-based predictive model, a conventional approach involves using a long short-term memory (LSTM) network for temporal feature extraction; however, transformers have demonstrated superior performance across a broad spectrum of applications \cite{zeyer2019comparison, karita2019comparative}. Initially designed to tackle bottlenecks in natural language processing \cite{vaswani2017attention}, they have since been found useful in many other domains, including vision \cite{carion2020end}.  Transformers have been shown to be better than LSTMs at identifying and exploiting long-range dependencies in sequential data.
Motivated by this, we explore the application of transformers in constructing a predictive model for more accurate anticipation of future events.

An essential consideration in the design of a planner involves the integration of uncertainty, which serves as an indicator of distribution shifts and the model's prediction confidence. The utilization of uncertainty as a concept has been extensively explored in various fields \cite{houlsby2011bayesian, chua2018deep}. Within the context of navigation, efforts have been made to incorporate uncertainty-aware exploration \cite{kahn2017uncertainty}. In the realm of balancing uncertainty and speed, a comparable strategy to ours can be found in RAMP \cite{sharma2023ramp}. 
This method utilizes LiDAR data and ground point inflation to map uncharted areas. It uses an MPPI-based planner to increase speed in low-uncertainty zones and decrease speed in high-uncertainty areas. Unlike RAMP, which prioritizes reducing speed in uncertainty, our approach optimizes the robot's trajectory to minimize uncertainty and it incorporates the world's dynamic into the planning.

\begin{figure}[t]
        \begin{center}
            \includegraphics[width=1\linewidth]{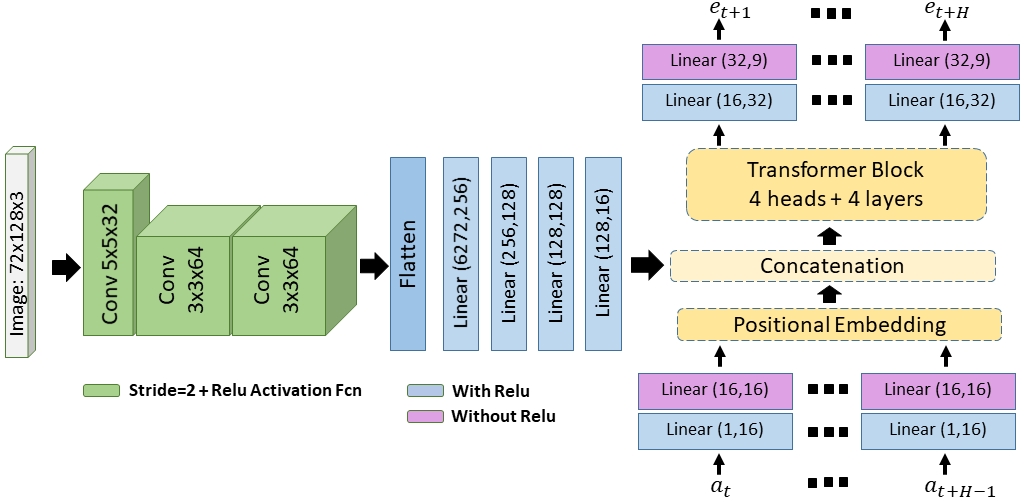}
        \end{center} 
        \caption{Our proposed transformer-based architecture.}
        \label{fig:architecture}
    \end{figure}
In a nutshell, we address mapless navigation challenges using a hybrid approach. We improve an environment predictive model with transformers and integrate uncertainty and nonlinear MPC for a hybrid planner. This approach optimizes robot navigation for speed, collision avoidance, and reduced uncertainty. The practical validation of our approach is conducted using real-world data gathered by an RC car, depicted in Fig. \ref{RC_car}. Subsequent sections delve into the methodology and present the results, accompanied by insightful observations. Ultimately, the conclusion section provides the culmination of our paper's findings.

\section{Methodology}
Our method is comprised of two key modules. The first is the image-based predictive model, which models the environment and leverages an RL approach suggesting a set of steering angles at each planning step. The second module is the MPC planner, which utilizes the vehicle's kinematic/dynamics model as well as the uncertainty in the environment's model to recommend throttle actions. In the subsequent sections, we provide an in-depth exploration of both modules. Besides, to train the image-based planner, we gathered a dataset and performed manual annotations to ensure its quality and have better evaluation studies. 

\subsection{Image-based Predictive Model}
The image-based planner operates by processing the current image alongside a designated set of actions covering ``p'' future steps, yielding the corresponding predictions for those upcoming events. This work adopts the model introduced in \cite{badgr} as the foundational reference. Utilizing these predicted events, a reward function is then formulated, enabling an online model-based RL approach. This involves the use of a global optimizer, such as the cross-entropy method~\cite{botev2013cross} or covariance matrix adaptation evolution strategy (CMA-ES) \cite{hansen2003reducing, hansen2019pycma}, to maximize the expected reward signal across time. Notably, re-planning occurs at each step, implying that only the initial action from the proposed set is executed. Existing models mainly use LSTM networks, but transformers have shown better performance across various applications. This study replaces LSTM modules with transformers, creating a new architecture as shown in Fig. \ref{fig:architecture}.
\begin{table}
     \centering
     \begin{tabular}{c|c|c|c}
        \hline
        Parameter & Unit & Description & Initial Value  \\
        \hline
        $C_1$ & - & Geometrical & 0.5 \\
        $C_2$ & $m^{-1}$ & Geometrical & 1.69 \\
        $C_{m_1}$ & $m/s^2$ & Motor Parameter & 12 \\
        $C_{m_2}$ & $1/s$ & Motor Parameter &  2.5 \\
        $C_{r_2}$ & $1/m$ & 2nd-order Friction Parameter & 0.15 \\
        $C_{r_0}$ & $m/s^2$ & 0th-order Friction Parameter & 0.7 \\
        $g$ & $m/s^2$ & Gravity & 9.81 \\
        \hline
    \end{tabular}   
     \caption{The model parameters.}
     \label{parameters_table}
\end{table}
 \begin{table*}[]
     \centering
     \begin{tabularx}{0.8875\textwidth}{| c | c | c | c | c | c | c | c | c | c | }
        \hline
        \textbf{Label} & Tree & Other Obstacles & Human & Waterhole & Mud & Jump & Traversable Grass & Smooth Road & Wet Leaves \\
        \hline
        \textbf{Samples} & 586 & 1631 & 517 & 66 & 267 & 164 & 6421 & 10632 & 698 \\
        \hline
    \end{tabularx}   
     \caption{Our datasets label distribution.}
     \label{tab:label_dist}
 \end{table*}

In order to estimate epistemic uncertainty we trained an ensemble of 5 image-based predictive models. This enables one to estimate the mutual information (MI), $I(\cdot,\cdot)$ between model outputs, $Z$, and weights, $W$, an established measure of epistemic uncertainty within the community \cite{houlsby2011bayesian, Berry_Meger_2023}. Our model adopts a multitask objective, assigning equal weight to the losses from both the image class prediction (classification task) and bearing estimation (regression task). For the classification task, we model an ensemble of categorical distributions, making it possible to derive the closed-form solution for $I(Z,W)$. However, when estimating epistemic uncertainty for the regression task, we create an ensemble of Gaussian distributions and $I(Z,W)$ no longer has a closed-form solution \cite{huber2008entropy}. To address this challenge, we rely on the pairwise-distance estimators (PaiDEs) \cite{kolchinsky2017estimating}. PaiDEs have been shown to efficiently estimate epistemic uncertainty for regression tasks for probabilistic neural networks by leveraging closed-form distributional distance between ensemble components \cite{berry2023escaping}. We utilize both Kullback-Liebler (KL) divergence and the Bhattacharyya distance between ensemble components for our PaiDEs.   
\subsection{MPC}
\begin{figure}[t]
        \begin{center}
            \includegraphics[width=0.9\linewidth]{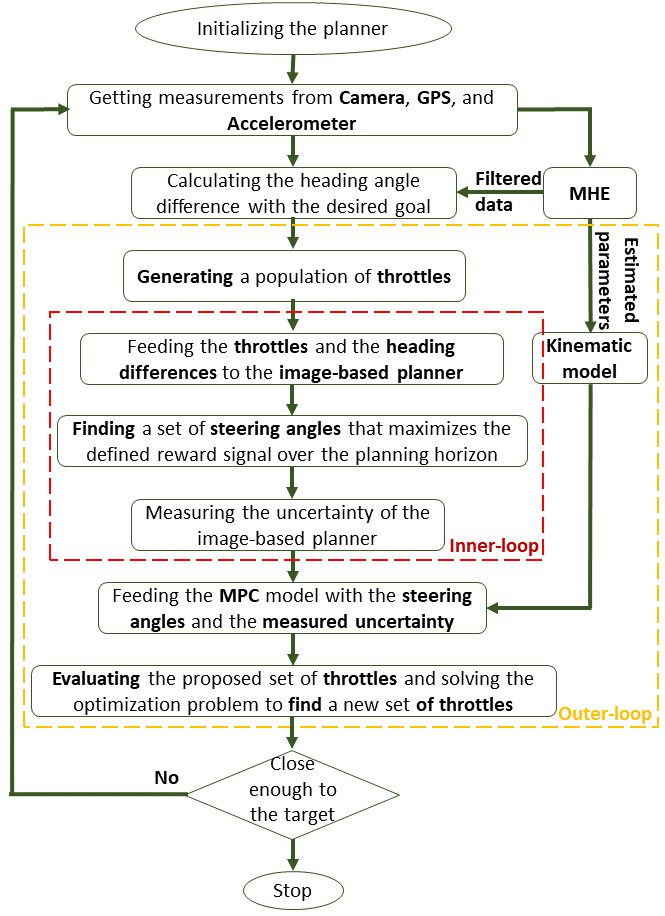}
        \end{center} 
        \caption{Flowchart of the proposed planner}
        \label{algorithm}
    \end{figure}

Regarding the MPC module, we have adopted the state-space model provided in~\cite{rcCar} and subsequently incorporated the slope effect into it:\\
\begin{eqnarray}
    &\dot{X} = V\cos(\psi + C_1\delta)\\ 
    &\dot{Y} = V\sin(\psi + C_1\delta)\\
    &\dot{\psi} = V\delta C_2\\ 
    &\dot{V} = (C_{m_1} - C_{m_2}V)D - C_{r_2}V^2 - C_{r_0} \nonumber \\ 
    &- (V\delta)^2 C_1{C_2}^2- Mg\sin(\phi)  \\ 
    &\dot{\phi} = 0 \\
    &\dot{\sigma} = prediction-model(\psi, D, input-image) 
    \label{vehicle_model}
\end{eqnarray}
The configuration and pose of the vehicle in GPS coordinates, its linear velocity, and the road slope are collectively represented by the state vector $[X, Y, \psi, V, \phi, \sigma] \in \Re^6$. The input vector, denoted as $[\delta, D]$, encapsulates both the steering angle and the throttle value. Furthermore, the model's parameters are detailed in Table. \ref{parameters_table}. To enhance the accuracy of the model during MPC iterations that assess various action sets, a key focus is placed on state and parameter estimation, aiming to minimize model uncertainty. To achieve this, we employ a moving horizon estimator \cite{lucia2017rapid}, which leverages a measurement window to estimate states and parameters. The measurements encompass GPS data, which includes the vehicle's position and orientation, alongside accelerometer data and camera data. While the standard model in the mentioned equation includes an additional uncertainty state, the dynamic nature of this uncertainty is encapsulated by the image-based module. Consequently, within the optimization process, MPC accommodates uncertainty, preventing undue speed increments irrespective of the world model's uncertainty. The flowchart of the proposed planner is presented in Fig. \ref{algorithm}. For MPC action set formulation and practical viability, we harness global optimization techniques as mentioned earlier, which eliminates the need for gradient flow in optimization, streamlining a potentially challenging and time-intensive aspect.

  \begin{table}[]
     \centering
     \begin{tabularx}{\columnwidth}{| c | X | X | X | c | }
        \hline
        \textbf{Model} & \textbf{Macro-Precision} & \textbf{Macro-Recall} & \textbf{Macro-F1-Score} & \textbf{Accuracy} \\
        \hline
        ResNet18 & 0.440 & 0.741 & 0.499 & 0.654 \\
        \hline
         ResNet50 & 0.627 & 0.838 & 0.701 & 0.808 \\
        \hline
         ResNet101 & 0.644 & 0.846 & 0.719 & 0.808 \\
         \hline
          VGG11 & \textbf{0.791} & 0.818 & 0.793 & 0.875 \\
        \hline
         VGG16 & 0.719 & 0.839 & 0.766 & 0.851 \\
        \hline
         ViT-B/16 & 0.685 & 0.844 & 0.749 & 0.854 \\
        \hline
        ViT-L/16 & 0.760 & \textbf{0.857} & \textbf{0.802} & \textbf{0.878} \\
        \hline

    \end{tabularx}   
     \caption{Results of the classification study on our datasets test set. }
     \label{tab:classification_results}
 \end{table}

\begin{figure}[t]
        \begin{center}
            \includegraphics[width=0.8\linewidth]{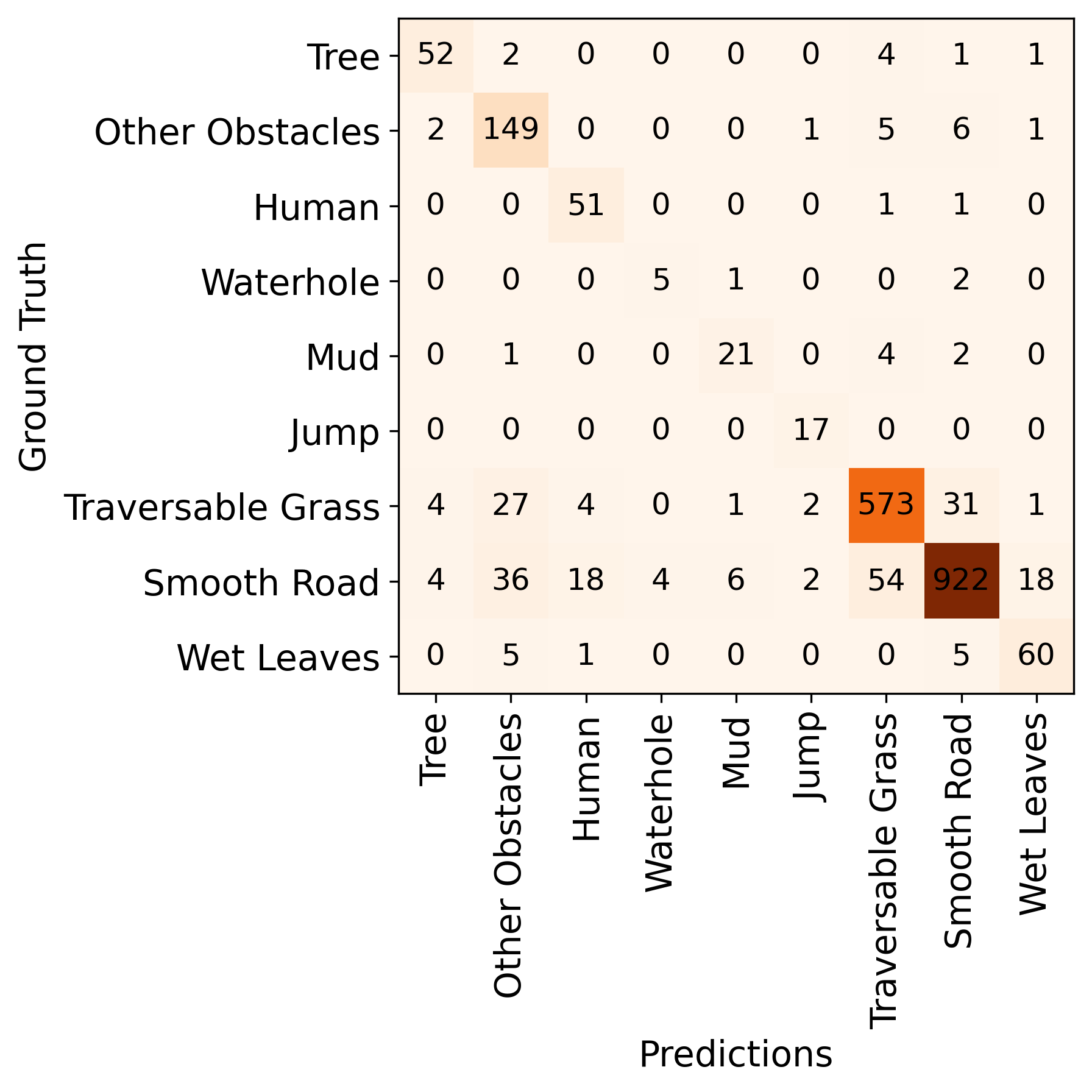}
        \end{center} 
        \caption{The ViT-L/16 model’s confusion matrix on our datasets test set.}
        \label{confusion_matrix}
    \end{figure}
\subsection{Dataset}
Our dataset encompasses trials of hands-on robot navigation, deliberately exposing the robot to intricate scenarios to enhance the dataset's diversity. This effort resulted in a carefully synchronized repository of 20,982 samples collected at a frequency of 5 Hz to minimize potential sample correlations. Unlike previous approaches, we opted for manual annotations to minimize potential labeling noise. Our annotations encompass nine distinct events, including ``tree'', ``other-obstacles'', ``human'', ``waterhole'', ``mud'', ``jump'', ``traversable-grass'', ``smooth-road'', and ``wet-leaves''. A detailed summary of the annotation distribution across our dataset is provided in Table. \ref{tab:label_dist}. To evaluate the complexity of our dataset on a classification task, we trained and evaluated three Residual Network (ResNet) \cite{he2016deep} models, two Vision Transformer (ViT) \cite{dosovitskiy2020image} models, and two Visual Geometry Group (VGG) \cite{simonyan2014very} models on our dataset. We trained each model using transfer learning and assessed their classification performance using established metrics such as macro-F1-score, macro-recall, and macro-precision. 

Table. \ref{tab:classification_results} presents the outcomes of our classification study, revealing that while the classification of our dataset proves challenging for ResNet18, it avoids excessive complexity that would necessitate a substantial alteration in results by augmenting the model's layers from ResNet50 to ResNet101. The confusion matrix depicted in Fig. \ref{confusion_matrix} demonstrates the Vision Transformer ViT-L/16 model's ability to effectively differentiate among the predefined categories. These results solidify the dataset's optimal level of intricacy, lending confidence to the feasibility of the classification task at hand. Lastly, preceding research restricted the scope of analyzed events, notably omitting labels for dynamic entities like humans. In contrast, our dataset meticulously annotates nine distinct events, opening avenues for comprehensive exploration across diverse applications.
    
\section{Results}

\begin{figure*}[]
        \begin{center}
            \includegraphics[width=0.9\linewidth]{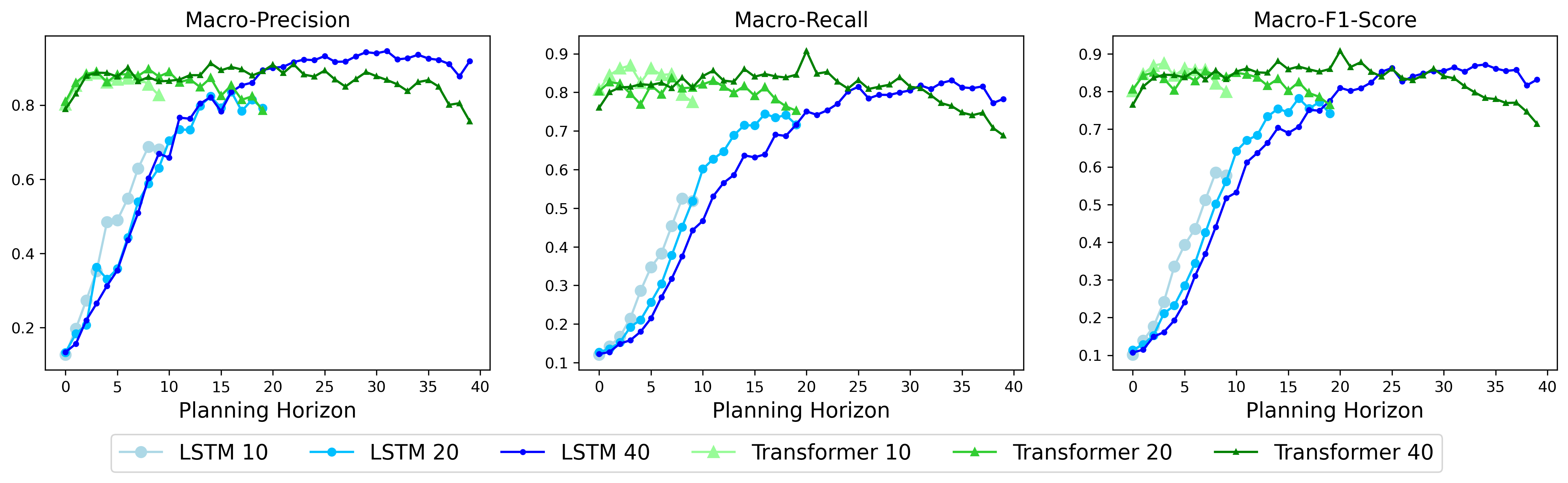}
        \end{center} 
        \caption{
    When assessing three distinct metrics, it becomes evident that the transformer-based model excels in terms of sustaining consistent performance throughout the planning horizon. Note that each model's evaluation is limited to the horizon it was originally trained for. }
        \label{fig_LSTM_result}
    \end{figure*}
\begin{figure*}[]
        \begin{center}
            \includegraphics[width=0.8\linewidth]{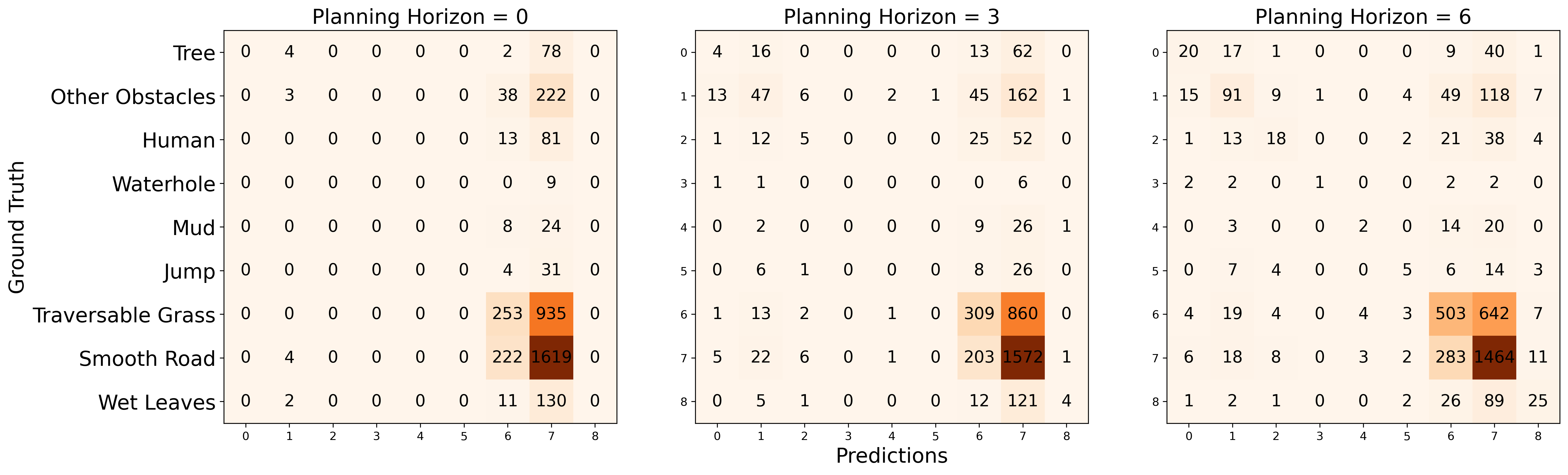}
        \end{center} 
        \caption{This figure depicts confusion matrices from a trained LSTM-based model with planning horizon of 40 at prediction steps $0$, $3$, and $6$. Initially, it mostly predicts "smooth-road" or "traversable-grass," but as the prediction horizon advances, the performance improves. Obstacles start to be recognized, and although some errors persist, the obstacles and smooth road become more distinguishable. Furthermore, we have 4,000 test samples, in contrast to the 2,000 samples shown in Fig. \ref{confusion_matrix}.}
        \label{LSTM_confusion_matrices}
\end{figure*}
\begin{figure*}[]
        \begin{center}
            \includegraphics[width=1\linewidth]{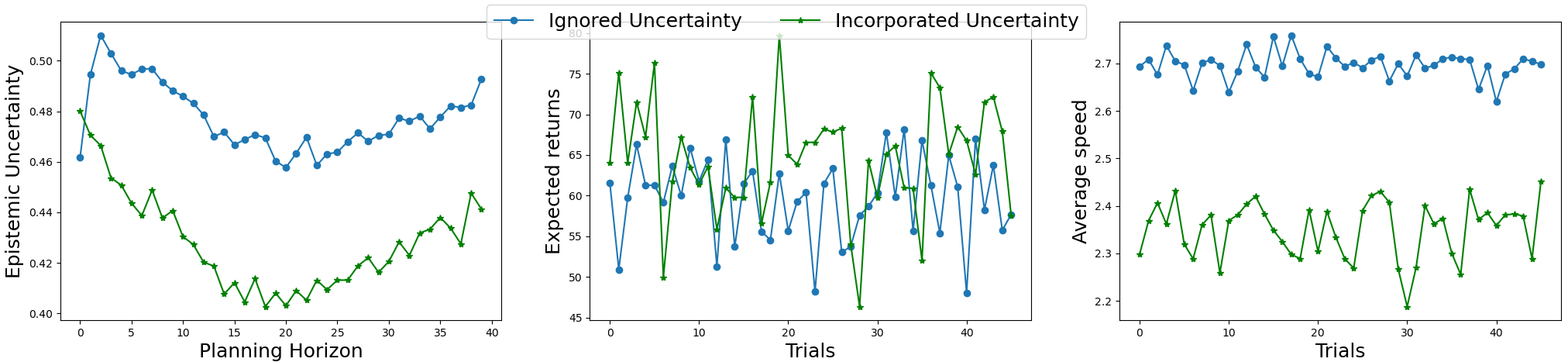}
        \end{center} 
        \caption{\andrew{The figure has a title integrated at the top.  It's not needed, due to the caption.} 
        MPC effectively mitigates epistemic uncertainty while also improving performance on favored terrains. Notably, the average of expected returns are $64.6$ and $59.7$ for the integrated and ignored uncertainty models, respectively.  \andrew{I don't think this figure shows that it maintains performance.  Also, which uncertainty measure is this?  KL?  Bhatt?  MI?  Is this for classification or regression?}}
        \label{final_study}
\end{figure*}
\subsection{Image-based predictive model}
In this section, we describe the studies we conducted to compare the LSTM and transformer-based techniques. Initially, we trained the pre-existing baseline model known as BADGR \cite{badgr} on our dataset. To delve deeper into the ramifications of the planning horizon, we train three models of each type with three different planning horizons: 10, 20, and 40 timesteps, and compare their performance. Please note that a horizon of 40 can be considered the ultimate planning limit, as it covers the next 8 seconds of planning. This duration is sufficient, given our maximum speed of $3 m/s$.  In every study, we conduct training for a total of 700 epochs, utilizing a batch size of 64, and employing an NVIDIA GeForce RTX 3090 GPU.  We evaluate each model on a test set including 4000 random samples, and the same random trajectories are used for all studies.  
\andrew{are the same random trajectories used in all three test sets?(yes)
}
\begin{figure}[]
        \begin{center}
            \includegraphics[width=0.7\linewidth]{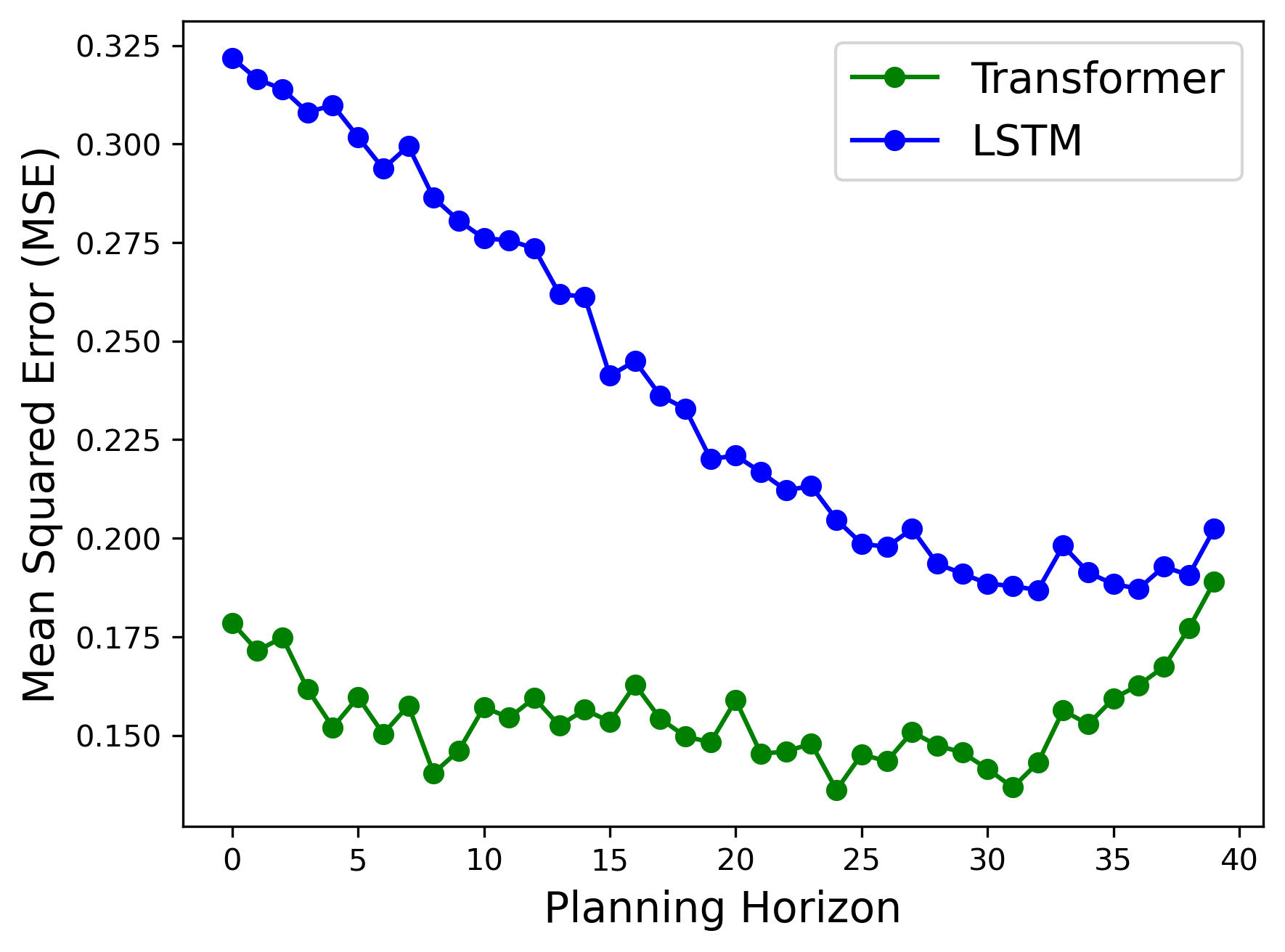}
        \end{center} 
        \caption{The results from the comparative study on bearing estimation errors further affirm the superior performance of the transformer-based model.}
        \label{fig_bearing}
\end{figure}

It's worth noting that this facet of inquiry has been largely overlooked in the existing literature, which has predominantly concentrated on the realm of RL, focusing on metrics such as expected returns. 

Fig. \ref{fig_LSTM_result} unveils the outcomes, highlighting a compelling ascending trend for the LSTM-based model as we progress through the prediction steps.
We think this pattern might arise from the LSTM networks' recurrent nature.\andrew{Are you sure about this?  How do you know?} As we progress in prediction, they extract increasingly valuable temporal features, enhancing predictive performance until reaching a saturation point. 
Moving on to the second study, we substituted the LSTM component, as delineated in the methodology section, with a transformer-based model, subsequently reiterating the study.

Fig. \ref{fig_LSTM_result} underscores a remarkable enhancement in the model's performance when using transformers. Notably, this improvement is most pronounced in its consistent prediction of future events. This trend continues in the domain of bearing estimation, as depicted in Fig. \ref{fig_bearing}, a pivotal component for guiding the vehicle toward the final goal.  The LSTM network has poor accuracy for the short-term predictions, but its accuracy increases for predictions farther in the future.  In contrast, the transformer-based model demonstrates impressive performance right from the start, although it experiences a slight decrease in predictive accuracy at later timesteps. 

The superiority of the transformer-based model over the LSTM lies in its capacity to effectively filter essential information through self-attention mechanisms. In contrast, LSTM networks, due to their recurrent nature, tend to exhibit biases towards future events, making them reliant on prior action inputs for accurate future event predictions. This key distinction underscores the advantage of the transformer-based model. To further probe the performance of the LSTM-based model, we conducted an additional study, increasing the number of time-steps between predictions to minimize correlations among future events. However, this adjustment did not alter the model's performance; the weakness in early predictions persisted. \andrew{I'd cut the reference to this study if you aren't reporting any data from it.}

This discrepancy holds great significance when the robot re-plans after each action. In this context, precise early-stage predictions become crucial, making the transformer-based architecture a preferable choice. To further assess the initial performance of LSTM, we present confusion matrices for the initial prediction steps in Fig. \ref{LSTM_confusion_matrices}. These matrices unveil the LSTM model's bias towards predicting ``smooth-road'' or ``traversable-grass''  in its early predictions.

\subsection{Uncertainty-aware hybrid planner}
In this section, we execute the complete pipeline, which incorporates the uncertainty module as well as the MPC with a specific objective in mind: to minimize both uncertainty and the time taken to achieve the goal at the same time.  We employ a strategy involving simulated higher speeds, achieved by augmenting the time steps in the planning process. 
This enhances input action diversity, improving the training and exploration of the uncertainty-aware planner. 
To elaborate, we designate a data sampling time 
of 0.2 seconds as the baseline speed, while 0.4 seconds equate to 2 times that speed (2X), and 0.6 seconds correspond to 3 times (3X) the baseline. Consequently, we adapt the training of the image-based component of the planner to align with these variations.

\begin{figure}[]
        \begin{center}
            \includegraphics[width=0.95\linewidth]{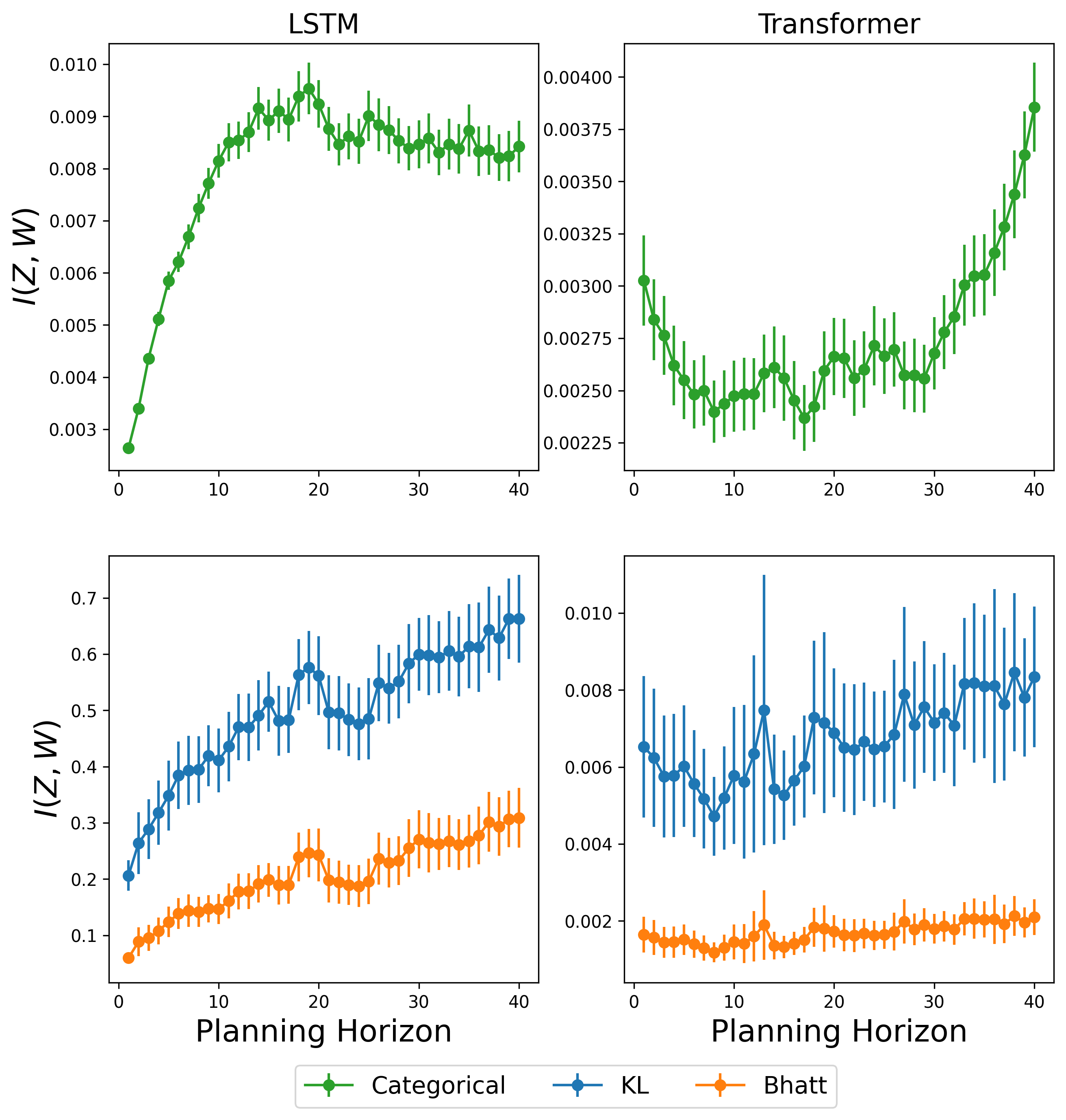}
        \end{center} 
        \caption{An exploration of uncertainty in LSTM and transformer-based models highlights that uncertainty grows along the planning horizon, results displayed with 95\% confidence interval. \andrew{Same as with the earlier version of figure 5, the problem here is that incomparable things (KL and Bhatt) are on the same plot, while the things we want to compare (LSTM and Transformer) are on different plots.}}
        \label{fig:uncertainty}
\end{figure}

Within this framework, the throttle range [0,1] is mapped to a speed range of [0, 3X baseline]. Following this, we train the predictive model with random speeds; we then compare the average uncertainty in each prediction step for both the LSTM and transformer based models. Fig. \ref{fig:uncertainty} presents the results, revealing that the uncertainty of both the LSTM and transformer models increases as the planning horizon increases for each task. This observation aligns with previous findings \cite{chua2018deep}, and is consistent with our intuition that there should be greater certainty for steps closer to the current state. Note that for the classification task, the $I(Z,W)$ is calculated directly. In contrast, for the regression task, we employed two pairwise-distance estimators, namely KL divergence and Bhattacharyaa distance (Bhatt), as outlined in \cite{berry2023escaping}.

Finally, we consider the transformer based architecture with the uncertainty incorporated, coupled with the MPC controller. This combination enables strategic planning for diverse scenarios using an RL approach, akin to the methodology outlined in previous works such as \cite{badgr, manderson2020-icra, wapnick2021}. For the image-based module, we adhere to the same reward function definition as: 
\begin{eqnarray}
& R({\hat{e}}_{t:t+H}^{0:K}) = - \sum_{t'=t}^{t+H-1} \hat{e}_{t'}^{coll} +  \\ & \alpha^{POS}R^{POS}(\hat{e}_{t'}^{0:K}) + \alpha^{BUM}R^{BUM}(\hat{e}_{t'}^{0:K}) \nonumber 
\label{common_reward}
\end{eqnarray}
Where $R^{POS} = (1 - \hat{e}_{t'}^{coll})\frac{1}{\pi}\angle(\hat{e}_{t'}^{POS}, p^{GOAL}) + \hat{e}_{t'}^{coll}$ and $R^{BUM} = (1 - \hat{e}_{t'}^{coll})\hat{e}_{t'}^{BUM} + \hat{e}_{t'}^{coll}$, with a total of $K$ events within a given horizon of $H$, and predicted probablities for each event represented as $\hat{e}$ e.g. $\hat{e}_{t'}^{coll}$ stands for the probability of having collision at time $t'$. These values are instrumental in directing the vehicle towards preferred terrains while ensuring collision avoidance. This function transforms the model's event predictions into rewards, and the RL agent then seeks to maximize the cumulative discounted reward over the defined horizon. 
In the MPC component, we tackle uncertainty by introducing a distinctive reward function, detailed as follows:
\begin{equation}
R^{MPC} = \frac{\beta_\sigma}{\sigma^2} + \beta_{V}V^2 
    \label{uncertainty_included}
\end{equation}
with $\beta_{\sigma}$ and $\beta_{V}$ as tuning parameters, and $\sigma$ representing the overall uncertainty. This helps to incorporate the imperative of uncertainty minimization. With this framework in place, we proceed to evaluate the effectiveness of our planner in real-world scenarios. The reason we can effectively balance the trade-off between speed and uncertainty using Eq. (\ref{uncertainty_included}) is due to the incorporation of the MPC algorithm. MPC proposes a set of throttle commands that inherently considers the kinematic model of the vehicle, allowing us to obtain critical vehicle's states information such as speed for each prediction step. Extracting this information directly from images is notably more challenging, although it is feasible, it could result in a complex and resource-intensive architecture.

Fig. \ref{final_study} summarizes our final study, which involved multiple trials. We used $\beta_{\sigma} = 10$ and $\beta_{V} = 1$ in our experiments. Our reward function optimizes throttle settings to minimize uncertainty, addressing various factors like input data distribution shifts. This doesn't always mean just using lower speeds during high uncertainty periods.
Given the transformer-based predictive model's consistent and promising performance, lower uncertainty suggests improved results when the input closely aligns with the training data. 
The prediction uncertainty curve shows a U-shaped pattern over prediction time, indicating increased confidence as more actions are considered. Initial uncertainty is high due to uncertainty about the robot's path, and it rises again at a prediction horizon threshold when current data is insufficient for accurate prediction. 
Finally, employing MPC in conjunction with the reward signal described in Eq. (\ref{uncertainty_included}) leads to lower average speeds and increased speed variance. This highlights how MPC aims to balance speed and uncertainty, making the overall planner more conservative in its use of maximum speeds.

\section{Conclusion}    
This paper proposes a hybrid planner that combines non-linear MPC and RL-based approaches, effectively addressing uncertainty. The RL algorithm employs an image-based predictive model to optimize steering angles based on reward maximization, while non-linear MPC recommends throttle settings that trade off goal-reaching time and uncertainty due to various reasons. Furthermore, we introduce a transformer-based architecture to model the environment. This model surpasses the current baseline by achieving consistent and applicable prediction accuracy across various prediction steps. Our dataset covering nine diverse events ensures rigorous evaluation, producing a balance between complexity and feasibility. Our results underscore the superiority of our proposed planner over the existing baseline, seamlessly integrating vehicle and environment models to generate viable, uncertainty-aware trajectories.

	{\small
		\bibliographystyle{ieeetr}
		\bibliography{egbib}

\begin{thebibliography}{10}

\bibitem{guastella2020learning}
D.~C. Guastella and G.~Muscato, ``Learning-based methods of perception and
  navigation for ground vehicles in unstructured environments: A review,'' {\em
  Sensors}, vol.~21, no.~1, p.~73, 2020.

\bibitem{islam2022off}
F.~Islam, M.~Nabi, and J.~E. Ball, ``Off-road detection analysis for autonomous
  ground vehicles: a review,'' {\em Sensors}, vol.~22, no.~21, p.~8463, 2022.

\bibitem{badgr}
G.~Kahn, P.~Abbeel, and S.~Levine, ``Badgr: An autonomous self-supervised
  learning-based navigation system,'' {\em IEEE Robotics and Automation
  Letters}, vol.~6, no.~2, pp.~1312--1319, 2021.

\bibitem{manderson2020-icra}
T.~Manderson, S.~Wapnick, D.~Meger, and G.~Dudek, ``Learning to drive off road
  on smooth terrain in unstructured environments using an on-board camera and
  sparse aerial images,'' in {\em 2020 IEEE International Conference on
  Robotics and Automation (ICRA)}, pp.~1263--1269, IEEE, 2020.

\bibitem{wapnick2021}
S.~Wapnick, T.~Manderson, D.~Meger, and G.~Dudek, ``Trajectory-constrained deep
  latent visual attention for improved local planning in presence of
  heterogeneous terrain,'' in {\em 2021 IEEE/RSJ International Conference on
  Intelligent Robots and Systems (IROS)}, pp.~460--467, IEEE, 2021.

\bibitem{garcia1989model}
C.~E. Garcia, D.~M. Prett, and M.~Morari, ``Model predictive control: Theory
  and practice—a survey,'' {\em Automatica}, vol.~25, no.~3, pp.~335--348,
  1989.

\bibitem{MPPI}
G.~Williams, P.~Drews, B.~Goldfain, J.~M. Rehg, and E.~A. Theodorou,
  ``Aggressive driving with model predictive path integral control,'' in {\em
  2016 IEEE International Conference on Robotics and Automation (ICRA)},
  pp.~1433--1440, IEEE, 2016.

\bibitem{yin2023risk}
J.~Yin, Z.~Zhang, and P.~Tsiotras, ``Risk-aware model predictive path integral
  control using conditional value-at-risk,'' in {\em 2023 IEEE International
  Conference on Robotics and Automation (ICRA)}, pp.~7937--7943, IEEE, 2023.

\bibitem{sharma2023ramp}
L.~Sharma, M.~Everett, D.~Lee, X.~Cai, P.~Osteen, and J.~P. How, ``Ramp: A
  risk-aware mapping and planning pipeline for fast off-road ground robot
  navigation,'' in {\em 2023 IEEE International Conference on Robotics and
  Automation (ICRA)}, pp.~5730--5736, IEEE, 2023.

\bibitem{zeyer2019comparison}
A.~Zeyer, P.~Bahar, K.~Irie, R.~Schl{\"u}ter, and H.~Ney, ``A comparison of
  transformer and lstm encoder decoder models for asr,'' in {\em 2019 IEEE
  Automatic Speech Recognition and Understanding Workshop (ASRU)}, pp.~8--15,
  IEEE, 2019.

\bibitem{karita2019comparative}
S.~Karita, N.~Chen, T.~Hayashi, T.~Hori, H.~Inaguma, Z.~Jiang, M.~Someki,
  N.~E.~Y. Soplin, R.~Yamamoto, X.~Wang, {\em et~al.}, ``A comparative study on
  transformer vs rnn in speech applications,'' in {\em 2019 IEEE Automatic
  Speech Recognition and Understanding Workshop (ASRU)}, pp.~449--456, IEEE,
  2019.

\bibitem{vaswani2017attention}
A.~Vaswani, N.~Shazeer, N.~Parmar, J.~Uszkoreit, L.~Jones, A.~N. Gomez,
  {\L}.~Kaiser, and I.~Polosukhin, ``Attention is all you need,'' {\em Advances
  in neural information processing systems}, vol.~30, 2017.

\bibitem{carion2020end}
N.~Carion, F.~Massa, G.~Synnaeve, N.~Usunier, A.~Kirillov, and S.~Zagoruyko,
  ``End-to-end object detection with transformers,'' in {\em European
  conference on computer vision}, pp.~213--229, Springer, 2020.

\bibitem{houlsby2011bayesian}
N.~Houlsby, F.~Husz{\'a}r, Z.~Ghahramani, and M.~Lengyel, ``Bayesian active
  learning for classification and preference learning,'' {\em arXiv preprint
  arXiv:1112.5745}, 2011.

\bibitem{chua2018deep}
K.~Chua, R.~Calandra, R.~McAllister, and S.~Levine, ``Deep reinforcement
  learning in a handful of trials using probabilistic dynamics models,'' {\em
  Advances in neural information processing systems}, vol.~31, 2018.

\bibitem{kahn2017uncertainty}
G.~Kahn, A.~Villaflor, V.~Pong, P.~Abbeel, and S.~Levine, ``Uncertainty-aware
  reinforcement learning for collision avoidance,'' {\em arXiv preprint
  arXiv:1702.01182}, 2017.

\bibitem{botev2013cross}
Z.~I. Botev, D.~P. Kroese, R.~Y. Rubinstein, and P.~L’Ecuyer, ``The
  cross-entropy method for optimization,'' in {\em Handbook of statistics},
  vol.~31, pp.~35--59, Elsevier, 2013.

\bibitem{hansen2003reducing}
N.~Hansen, S.~D. M{\"u}ller, and P.~Koumoutsakos, ``Reducing the time
  complexity of the derandomized evolution strategy with covariance matrix
  adaptation (cma-es),'' {\em Evolutionary computation}, vol.~11, no.~1,
  pp.~1--18, 2003.

\bibitem{hansen2019pycma}
N.~Hansen, Y.~Akimoto, and P.~Baudis, ``{CMA-ES/pycma} on {G}ithub.'' Zenodo,
  DOI:10.5281/zenodo.2559634, Feb. 2019.

\bibitem{Berry_Meger_2023}
L.~Berry and D.~Meger, ``Normalizing flow ensembles for rich aleatoric and
  epistemic uncertainty modeling,'' {\em Proceedings of the AAAI Conference on
  Artificial Intelligence}, vol.~37, no.~6, pp.~6806--6814, 2023.

\bibitem{huber2008entropy}
M.~F. Huber, T.~Bailey, H.~Durrant-Whyte, and U.~D. Hanebeck, ``On entropy
  approximation for gaussian mixture random vectors,'' in {\em 2008 IEEE
  International Conference on Multisensor Fusion and Integration for
  Intelligent Systems}, pp.~181--188, IEEE, 2008.

\bibitem{kolchinsky2017estimating}
A.~Kolchinsky and B.~D. Tracey, ``Estimating mixture entropy with pairwise
  distances,'' {\em Entropy}, vol.~19, no.~7, p.~361, 2017.

\bibitem{berry2023escaping}
L.~Berry and D.~Meger, ``Escaping the sample trap: Fast and accurate epistemic
  uncertainty estimation with pairwise-distance estimators,'' {\em arXiv
  preprint arXiv:2308.13498}, 2023.

\bibitem{rcCar}
R.~Verschueren, S.~De~Bruyne, M.~Zanon, J.~V. Frasch, and M.~Diehl, ``Towards
  time-optimal race car driving using nonlinear mpc in real-time,'' in {\em
  53rd IEEE conference on decision and control}, pp.~2505--2510, IEEE, 2014.

\bibitem{lucia2017rapid}
S.~Lucia, A.~T{\u{a}}tulea-Codrean, C.~Schoppmeyer, and S.~Engell, ``Rapid
  development of modular and sustainable nonlinear model predictive control
  solutions,'' {\em Control Engineering Practice}, vol.~60, pp.~51--62, 2017.

\bibitem{he2016deep}
K.~He, X.~Zhang, S.~Ren, and J.~Sun, ``Deep residual learning for image
  recognition,'' in {\em Proceedings of the IEEE conference on computer vision
  and pattern recognition}, pp.~770--778, 2016.

\bibitem{dosovitskiy2020image}
A.~Dosovitskiy, L.~Beyer, A.~Kolesnikov, D.~Weissenborn, X.~Zhai,
  T.~Unterthiner, M.~Dehghani, M.~Minderer, G.~Heigold, S.~Gelly, {\em et~al.},
  ``An image is worth 16x16 words: Transformers for image recognition at
  scale,'' {\em arXiv preprint arXiv:2010.11929}, 2020.

\bibitem{simonyan2014very}
K.~Simonyan and A.~Zisserman, ``Very deep convolutional networks for
  large-scale image recognition,'' {\em arXiv preprint arXiv:1409.1556}, 2014.

\end{thebibliography}
	}

\end{document}